\documentclass[10pt,twocolumn,letterpaper]{article}
\pdfoutput=1 % arxiv

\usepackage{cvpr}
\usepackage{times}
\usepackage{epsfig}
\usepackage{graphicx}
\usepackage{amsmath}
\usepackage{amssymb}

\usepackage{amsfonts}
\usepackage{physics} 
\usepackage[protrusion=true,tracking=true,kerning=true,expansion,final]{microtype}      % microtypography
\usepackage{xcolor} %colors
\usepackage{booktabs} % for professional tables
\usepackage{subcaption}
\usepackage{nicefrac} % nice flat fraction
\usepackage{cancel}
\usepackage[square,sort,comma,numbers]{natbib}
\usepackage{xspace}

\usepackage{gensymb}
\usepackage{bbding}
\usepackage{mathtools}
\usepackage{centernot} % not parallel, etc.
\usepackage{bigints}
\usepackage{dsfont} %mathbb 1
\usepackage{esint} % beatiful integrals
\usepackage{amsthm} % theorems
\usepackage{authblk}
\usepackage{multirow}

\usepackage{varioref} % references to show page
\usepackage[pagebackref=true,breaklinks=true,letterpaper=true,colorlinks,bookmarks=false]{hyperref}
\usepackage[capitalise]{cleveref} 
\crefname{appsec}{appendix}{appendices}
\Crefname{appsec}{Appendix}{Appendices}

\usepackage{url}

\definecolor{mydarkblue}{rgb}{0,0.08,0.45}
\definecolor{urlcolor}{rgb}{0,.145,.698}
\definecolor{linkcolor}{rgb}{.71,0.21,0.01}
\hypersetup{ %
	pdftitle={Colored Noise Injection for Training Adversarially Robust Neural Networks}, % TODO
	pdfauthor={Evgenii Zheltonozhskii, Chaim Baskin, Yaniv Nemcovsky, Brian Chmiel, Alex M. Bronstein, and Avi Mendelson},
	pdfsubject={},
	pdfkeywords={},
	pdfborder=0 0 0,
	pdfpagemode=UseNone,
	breaklinks=true,  % so long urls are correctly broken across lines
	colorlinks=true,
	letterpaper=true,
	bookmarks=false,
	urlcolor=urlcolor,
	linkcolor=linkcolor,
	citecolor=mydarkblue,
	filecolor=mydarkblue,
	pdfview=FitH}

\renewcommand*{\backref}[1]{} % for backref < 1.33 necessary
\renewcommand*{\backrefalt}[4]{%
	\ifcase #1 %
	\or
	(cited on p. #2)%
	\else
	(cited on pp. #2)%
	\fi
}

\vbadness=\maxdimen
\usepackage{chngcntr}
\usepackage{apptools}
\AtAppendix{\counterwithin{figure}{section}}
\AtAppendix{\counterwithin{table}{section}}

\usepackage{docmute} 
\usepackage{xr}
\cvprfinalcopy % *** Uncomment this line for the final submission

\def\cvprPaperID{34} % *** Enter the CVPR Paper ID here

\usepackage{bm}

\ifcvprfinal\fi

\title{Colored Noise Injection for Training Adversarially Robust Neural Networks}

\newcommand*\samethanks[1][\value{footnote}]{\footnotemark[#1]}

\author[1]{Evgenii Zheltonozhskii\thanks{Equal contribution.}}
\author[1]{Chaim Baskin\samethanks[1]}
\author[1]{Yaniv Nemcovsky}
\author[1,2]{Brian Chmiel}
\author[1]{Avi Mendelson}
\author[1]{Alex M. Bronstein}
\affil[1]{Technion – Israel Institute of Technology}
\affil[2]{Intel – Artificial Intelligence Products Group (AIPG)}
\affil[ ]{\tt\small
\href{mailto:evgeniizh@campus.technion.ac.il}{evgeniizh@campus.technion.ac.il};
\href{mailto:chaimbaskin@cs.technion.ac.il}{chaimbaskin@cs.technion.ac.il};
\href{mailto:yanemcovsky@cs.technion.ac.il}{yanemcovsky@cs.technion.ac.il};
\href{mailto:brian.chmiel@intel.com}{brian.chmiel@intel.com};
\href{mailto:avi.mendelson@cs.technion.ac.il}{avi.mendelson@cs.technion.ac.il}
\href{mailto:bron@cs.technion.ac.il}{bron@cs.technion.ac.il};}

\begin{document}

\maketitle
%\thispagestyle{empty}
 
%%%%%%%%% ABSTRACT
\begin{abstract}
Even though deep learning has shown unmatched performance on various tasks, neural networks have been shown to be vulnerable to small adversarial perturbations of the input that lead to significant performance degradation. In this work we extend the idea  of adding white Gaussian noise to the network weights and activations during adversarial training (PNI \cite{rakin2018parametric})  to the injection of colored noise for defense against common white-box and black-box attacks.
We show that our approach outperforms PNI and various previous approaches in terms of adversarial accuracy on CIFAR-10 and CIFAR-100 datasets. 
In addition, we provide an extensive ablation study of the proposed method justifying the chosen configurations.

%A \href{https://github.com/yanemcovsky/SIAM}{reference implementation} of the proposed techniques is provided. % in the supplementary material.  
\end{abstract}

%%%%%%%%% BODY TEXT

\section{Introduction}
\label{sec:intro}
Deep Neural Networks (DNNs) have shown a tremendous success in a variety of applications, including image classification and generation, text recognition,  machine translation, playing games, etc.
Despite achieving notable performance on numerous tasks, DNNs appear to be sensitive to small perturbations of the inputs. \citet{szegedy2013intriguing} have shown that 
it is possible to exploit this sensitivity to create \emph{adversarial examples} -- visually indistinguishable inputs which are classified differently. Subsequent studies proposed different adversarial attacks --- techniques for creating adversarial examples. 

 One of the first practical attacks is FGSM \cite{goodfellow2014explaining}, which used the appropriately scaled sign of the attacked network's gradient. PGD \cite{madry2018towards}, one of the strongest attacks to date, improved FGSM by repeating the gradient step iteratively, i.e., performing projected gradient ascent in the neighbourhood of the input. C\&W \cite{carlini2017towards} used a loss term penalizing large distances from the orginal input instead of applying hard restriction on it. In this way, the resulting attack is \emph{unbounded}, i.e., tries to find a minimum norm adversarial example rather than searching for it in predefined region. DDN \cite{rony2019decoupling} significantly improved the runtime and performance of C\&W  by  decoupling optimization of the direction and the norm.

It was noted that it is possible to create adversarial examples even without access to internals of the model and, in particular, its gradients, i.e., treating the model as a black box (as opposed to previously mentioned white box attacks). The approaches to black box attacks can be roughly divided into two main classes of approaches: the first class trains a different model with known gradients to generate adversarial examples and then transfer them to the victim model \cite{papernot2017practical,liu2016delving}. The second class attempts to estimate gradients of the model numerically, based solely on its inputs and outputs \cite{chen2017zoo,li2019nattack,wierstra2008natural,anonymous2020bayesopt}.

In order to confront with adversarial attacks, it was suggested to add the adversarial examples to the training process and balance between them and the original images \cite{szegedy2013intriguing,madry2018towards}. Many subsequent works have tried to increase the strength of training-time attacks to improve robustness \cite{khoury2019adversarial,liu2019training,jiang2018learning,balaji2019instance,zantedeschi2017efficient}.
A different approach to overcome adversarial attacks is to add randomization to the neural network \cite{zheng2016improving,zhang2019defending}, making it harder for the attacker to evaluate the gradients and thus to exploit the vulnerability of the network. 
Recently, \citet{rakin2018parametric} proposed to add Gaussian noise to the weights and activation of the network and showed improvement over "vanilla" adversarial training under various attacks. 

%\paragraph{Contribution.}  
In this paper, we propose a generalization of parametric noise injection (PNI) \cite{rakin2018parametric} which we henceforth term parametric \emph{colored} noise injection (CNI). The main idea is to replace the independent noise with low-rank multivariate Gaussian noise. We show that this modification provides consistent accuracy improvement under various attacks on a number of datasets.
 
\section{Method}
\label{sec:method}
% TODo extend on motivation 
In this section we introduce the proposed method of colored noise injection for adversarial defence (CNI). %We will emphasize the difference between our method and previous proposed PNI.
The previously proposed PNI \cite{rakin2018parametric} has much in common with uncorrelated variational dropout \cite{kingma2015variational}, a powerful regularization technique.  In both methods, the noise is distributed as:
\begin{equation}
    \epsilon \sim \mathcal{N}(0, \Lambda),
\end{equation}
for a diagonal $N\times N$  matrix $\Lambda = \mathrm{Diag}(\lambda)$. Both methods optimize the parameters $\lambda$ during training. The difference between two methods lies in their objective: while varational dropout attempts to infer the Bayesian posterior, PNI makes use of adversarial training to optimize the trade-off between the clean (unperturbed) and adversarial accuracy. In the adversarial training scheme, lowering the noise strength minimizes the clean loss, while increasing the strength provides a defense from adversarial attacks, thereby minimizing the adversarial loss.

\citet{kingma2015variational} have studied the addition of both correlated and uncorrelated random noise to the weights, claiming dropout \cite{srivastava2014dropout} is a particular case of such additive noise. Specifically, the advantage of correlated noise over uncorrelated one was demonstrated. Nonetheless, \citet{rakin2018parametric} have only considered the addition of uncorrelated noise. We therefore consider a generalization of PNI which is based on colored (correlated) noise. We model such noise using the multivariate normal distribution with a low-rank covariance. For an $N$-dimensional noise vector, the noise with an $M$-ran covarianceis distributed as
\begin{equation}
    \epsilon \sim \mathcal{N}(0, \Sigma),
\end{equation}
where
\begin{equation}
    \Sigma = \Lambda + VV^\top,
\end{equation}
where  $\Lambda$ is an $N\times N$ non-negative diagonal matrix, and $V$ is an  $N\times M$ matrix. Note that PNI is a particular case of CNI with $M=0$. The off-diagonal part of the covariance matrix, $VV^\top$, is a general positive semi-definite symmetric matrix with the rank upper-bounded by $M$ representing a low-dimensional interaction between different parameters.

\paragraph{Sampling low-rank multivariate normal noise}
In order to sample the noise, we make use of the decomposition of the covariance matrix  $\Sigma = \Lambda + VV^\top$. We sample two independent normal vectors,
\begin{align}
    \epsilon_{D}, \epsilon_{C} &\sim \mathcal{N}(0,I_N),
\end{align}
and let
\begin{align}
    \epsilon &= \Lambda^{\nicefrac{1}{2}}\epsilon_D +  V\epsilon_C.
\end{align}

% We tried to improve the quality of defence by introduction of correlation between different components of noise vector injected into weights or activations. 
%We denote one with colored noise as CPNI (colored PNI).

\paragraph{Weight decay}
We noted that for WideResNet  the noise strength increases significantly as compared to ResNet. This leads to very slow convergence and lower performance of the resulting model. To overcome this phenomenon, we added an additional weight decay term to the elements of $V$. While this approach leads to faster covergence and competitive results on both clean and adversarial datasets, it introduces an additional hyperparameter that requires some tuning.
\section{Experiments}
\label{sec:exp}

\begin{table}
    \centering
    \caption{Comparison of our method (CNI) to PNI \cite{rakin2018parametric} using various configurations on CIFAR-10 with ResNet20 under PGD attack with $k=7$ iterations. Mean and standard deviation are calculated over $10$ runs  in our experiments (upper half), and over $5$ runs in the experiments by \citet{rakin2018parametric}  (lower half). Noise is injected either to the weights (``W'') or the output activations (``A-a''). Best results for PNI and CNI are set in {\bf bold}.}
    \begin{tabular}{@{}lcc@{}} 
        \toprule
        \multirow{2}{*}{\textbf{Method}} &\multicolumn{2}{c}{\textbf{Accuracy,  mean$\bm{\pm}$std\%} } \\ 
        \cmidrule(r){2-3}
         & \textbf{Clean} & \textbf{PGD} \\ 
        \midrule
        Adversarial training\cite{madry2018towards}  & $83.84\pm0.05$    & $39.14\pm0.05$ \\  
        PNI-W  & $82.84\pm0.22$  & $\mathbf{46.11\pm0.43}$  \\
        CNI-W   & $78.48\pm0.41$    & $\mathbf{48.84\pm0.55}$  \\
        CNI-A-a   & $\mathbf{83.41\pm0.14}$    & $45.47\pm0.18$  \\
        CNI-W+A-a   & $77.07\pm0.40$    & $46.07\pm0.45$  \\
        \midrule
        PNI-W  & $84.89\pm0.11$   & $45.94\pm0.11$  \\
        PNI-W+A-a   & $\mathbf{85.12\pm0.10}$    & $43.57\pm0.12$  \\
        \bottomrule
    \end{tabular}
    \label{tab:results_cpni}
\end{table}

\begin{table}
    \centering
    \caption{
    Comparison of our method to prior art  against black-box attacks on CIFAR-10, ResNet-20 under transferable PGD attack and NAttack \cite{li2019nattack}. $^\dagger$ denotes our evaluation of the code provided by authors or our re-implementation thereof.  
     }
    \centering
       
    \begin{tabular}{@{}ccc@{}} 
        \toprule
        \multirow{2}{*}{\textbf{Method}}   &\multicolumn{2}{c}{\textbf{Accuracy,\%}} \\ 
        \cmidrule(r){2-3}
        & \textbf{Transferable attack} & \textbf{NAttack} \\ 
        \midrule
         Adv. training \cite{madry2018towards} $^\dagger$ &  $\mathbf{58.8}$ & $33.17 $  \\
          PNI-W \cite{rakin2018parametric} $^\dagger$ &$54.6$  & $47.17 $ \\
          CNI-W (our) &$54.1$& $\mathbf{48.91} $  \\
         % 16 & 0.0 & $\pm$  & $\pm$  & $\pm$ \\
        \bottomrule
    \end{tabular}
   
    \label{tab:results_bb_cf}
    
    \hfill

\end{table}

\begin{table}
    \centering
    \caption{
    Comparison of our method to prior art on CIFAR-10 with WideResNet-28-4 under PGD attack with $k=10$. Mean and standard deviation are calculated over  $10$ runs  in our experiments, and over $2$ runs in MMA. $^\dagger$ denotes our evaluation of the code provided by authors or our re-implementation thereof.  $^+$ denotes our evaluation based on the checkpoint provided by the authors. We also provide results for a larger (WideResNet-34-10) network in the lower part of the table. 
     }
    \centering
      
        \begin{tabular}{@{}lcc@{}} 
        \toprule
        \multirow{2}{*}{\textbf{Method}}&\multicolumn{2}{c}{\textbf{Accuracy,  mean$\bm{\pm}$std\%}  } \\ 
        \cmidrule(r){2-3}
         & \textbf{Clean} & \textbf{PGD}  \\ 
        \midrule
        Adv. training \cite{madry2018towards}$^\dagger$    & $86.08\pm0.00$    & $38.58\pm0.06$ \\        
%        L2L \cite{jiang2018learning}   & $85.31\pm0.41$ & $53.42\pm 1.07$  \\
        MMA \cite{Ding2020MMA}  &$\mathbf{86.24\pm0.13}$ & $54.86\pm1.16$  \\
        PNI \cite{rakin2018parametric}$^\dagger$   & $84.63\pm 0.15$   & $53.34\pm0.29$ \\
        CNI-W (our)  & $84.42\pm 0.28$    & $\mathbf{55.76\pm 0.29}$  \\ \midrule
    %    CSAT \cite{sarkar2019enforcing} & $87.65$ & $54.77$ \\
        IAAT \cite{balaji2019instance} &$\mathbf{91.3}$ & $48.53$  \\
        TRADES \cite{DBLP:conf/icml/ZhangYJXGJ19}$^{+}$ & $84.92$  &$56.5$   \\
        MART \cite{DBLP:conf/icml/ZhangYJXGJ19}$^{+}$ & $83.62$  &$\mathbf{57.3}$   \\
   %     Dynamic adversarial training\cite{pmlr-v97-wang19i}& $84.51$  &$55.03$   \\
%        Bilateral Adv Training\cite{wang2019bilateral}& $91.00$ &$57.5$  \\
        \bottomrule
    \end{tabular} 
    \label{tab:comp_cifar10_wide}

\end{table}
 
\begin{table}
    \centering
    \caption{
    Comparison of our method to prior art on CIFAR-100 with WideResNet-28-4 under PGD attack with $k=10$.  Mean and standard deviation is calculated over  10 runs. $^\dagger$ denotes  our evaluation of the code provided by authors or our re-implementation thereof.  $^+$ denotes our evaluation based on the checkpoint provided by the authors. We also provide results for a larger (WideResNet-34-10) network in the lower part of the table. 
     }
    \begin{tabular}{@{}lcc@{}} 
        \toprule
        \multirow{2}{*}{\textbf{Method}}&\multicolumn{2}{c}{\textbf{Accuracy,  mean$\bm{\pm}$std\%}  } \\ 
        \cmidrule(r){2-3}
         & \textbf{Clean} & \textbf{PGD}  \\ 
        \midrule
         Adv. training \cite{madry2018towards}$^\dagger$  &  $\mathbf{56.60\pm0.00}$  & $ 16.71\pm0.04$ \\
       
        PNI \cite{rakin2018parametric}$^\dagger$   & $54.03\pm0.37$   & $23.67\pm 0.33$  \\
        CNI-W (our)  & $53.65\pm 0.25$    & $\mathbf{25.03\pm 0.21}$  \\
        \midrule
        IAAT \cite{balaji2019instance}  & $\mathbf{68.1}$ & $26.17$  \\ L2L \cite{jiang2018learning}  & $ 60.95 \pm 0.13$   & $\mathbf{31.03  \pm 0.50}$ \\
        \bottomrule
    \end{tabular} 
     \label{tab:comp_cifar100_wide}
\end{table}

\paragraph{Experimental settings.}
We trained ResNet-20 defended with CNI for 400 epochs on CIFAR-10 using SGD with the learning rate $0.1$, reduced by $10$ at epochs 200 and 300, and weight decay $10^{-4}$. The number of iteration of the PGD for adversarial training was set to $k=7$.
In our experiments we chose rank $M=5$ for the colored noise factor. In \cref{sub:ablation} we show the effect of different values of $M$ to the final accuracy. We have not studied other distributions except the multivariate normal.

For WideResNet-28-4, we used the Ranger optimizer (RAdam \cite{liu2020radam} with lookahead \cite{zhang2019lookahead}) for 100 epochs,  with the learning rate $0.1$, reduced by $10$ at epochs 75 and 90, weight decay $10^{-4}$ and additional weight decay of $10^{-3}$ for CIFAR-10 and $3\cdot 10^{-3}$ for CIFAR-100 for the elements of $V$. The number of iteration of the PGD for adversarial training was set to $k=10$.
In all cases, the model with highest accuracy on a clean validation set was chosen for the evaluation.

\paragraph{White box attacks.}
We evaluate our defense against PGD attack \cite{madry2018towards} with same number of iterations as used in adversarial training ($k=7$ for ResNet-20 and $k=10$ for WideResNet-28-4). The results are reported in \cref{tab:results_cpni,tab:comp_cifar10_wide,tab:comp_cifar100_wide}. CNI outperforms other methods using WideResNet-28-4 and shows compatible results even when compared with methods which use larger networks.
%In addition, we evaluate a median radius of succesful C\&W $L_2$ \cite{carlini2017towards} attack as a measure of the robustness.

\paragraph{Black box attacks.}
We evaluated the proposed method against two common black-box attacks, in particular, the transferable attack  \cite{liu2016delving} and NAttack\cite{li2019nattack}. For the transferable attack, we trained another instance of the CNI-W model and used it as a source model in two configurations: PGD with and without smoothing. 
The results are reported in \cref{tab:results_bb_cf}. For the transferable attack, our method achieve comparable results to previous art and outperforms them on  NAttack.

\subsection{Ablation study}
\label{sub:ablation}
\begin{figure}
\centering
\includegraphics[width=\linewidth]{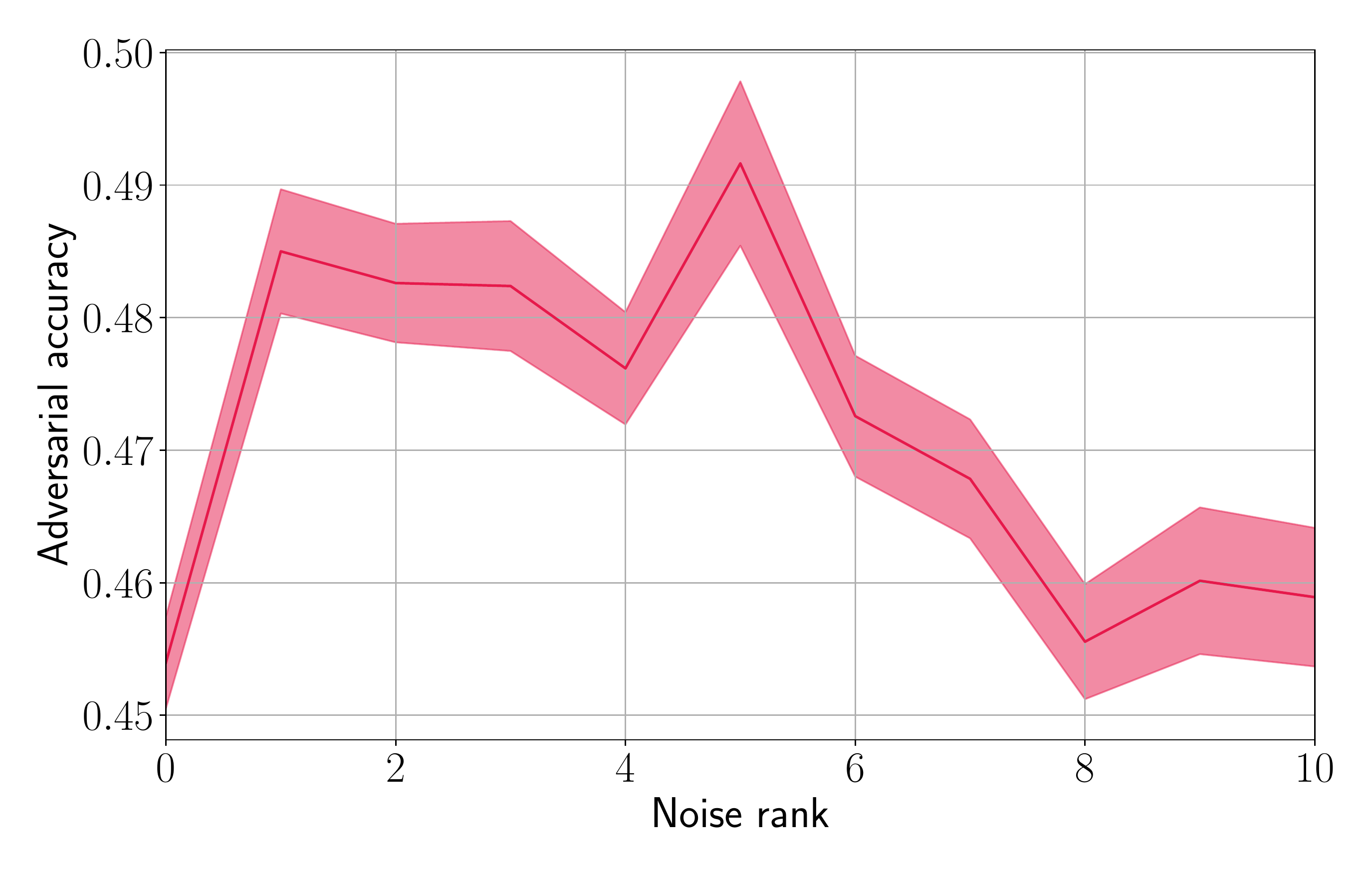}
\caption{Accuracy of CNI-W model under PGD attack with different noise covariance rank. Shaded region shows standard deviation of the results calculated over 50 runs.}
\label{fig:rank_ablation}
\end{figure}

We study the dependence of the network performance on noise rank. The results are shown in \cref{fig:rank_ablation}. As we can see, coloring the noise gives significant improvement of adversarial accuracy, while too high rank of the noise reduces the accuracy, probably due to overparametrization.

\begin{figure}
\centering
\includegraphics[width=\linewidth]{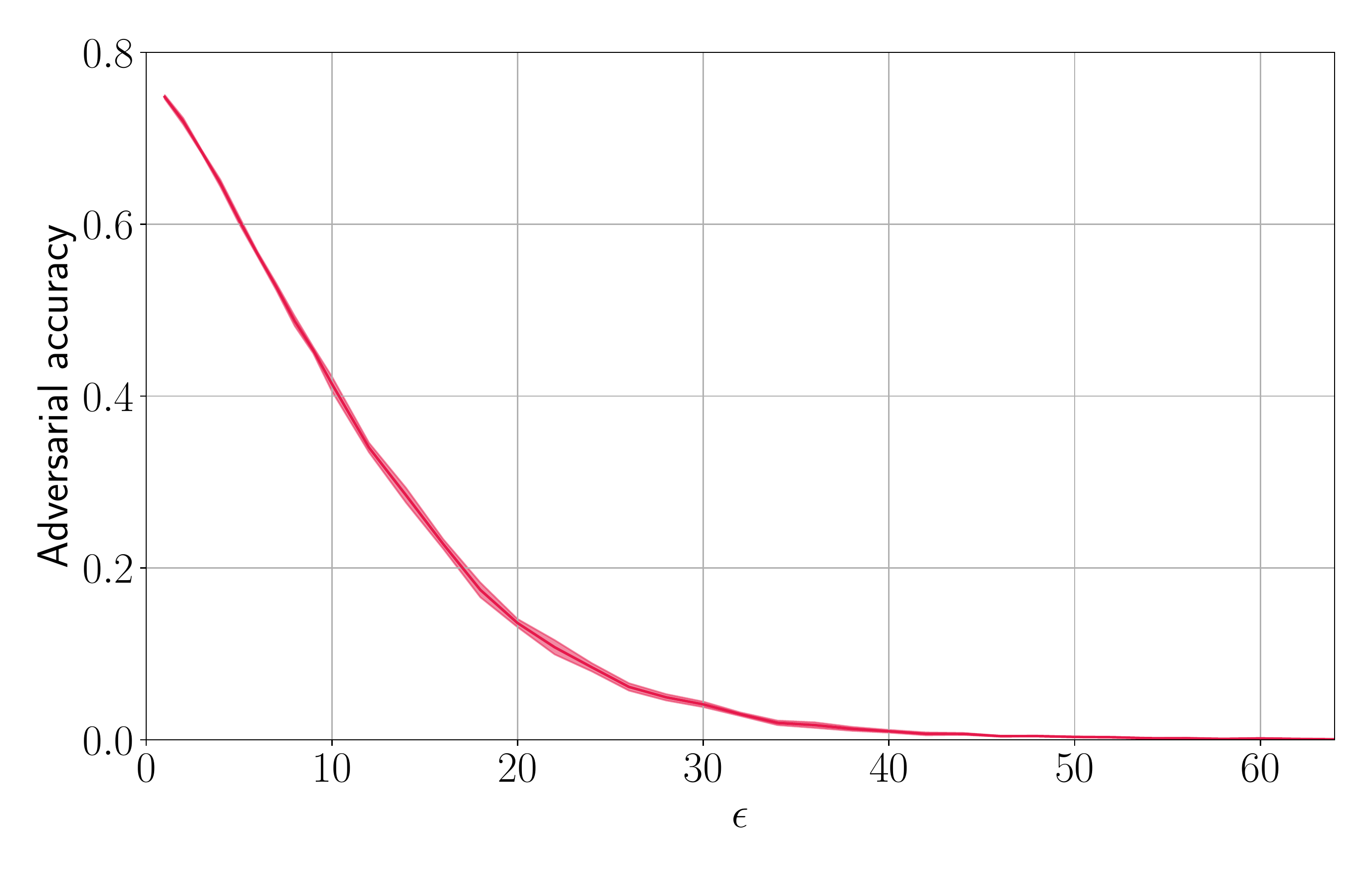}
\caption{Accuracy of CNI-W model under PGD attack with different attack radius, $\epsilon$ (255 scale). Shaded region shows standard deviation of the results calculated over 5 runs.}
\label{fig:eps_ablation}
\end{figure}

\begin{figure}
\centering
\includegraphics[width=\linewidth]{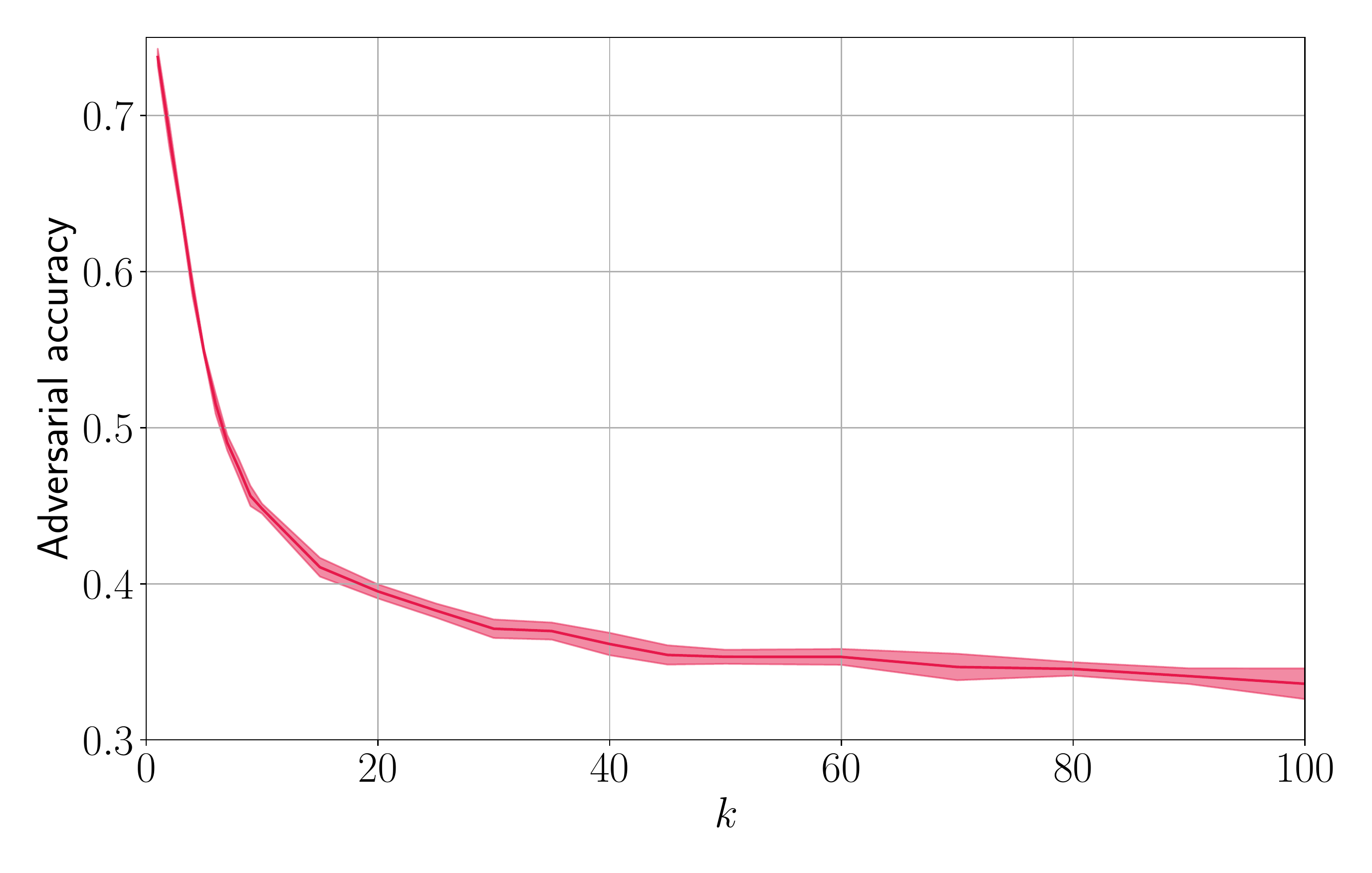}
\caption{Accuracy of CNI-W model under PGD attack with different number of iterations, $k$. Shaded region shows standard deviation of the results calculated over 5 runs.}
\label{fig:k_ablation}
\end{figure}

We also study the adversarial accuracy as a function of $\epsilon$ and $k$ (\cref{fig:eps_ablation,fig:k_ablation}). \cref{fig:k_ablation} shows that CNI has relatively low variance of the results at small number of iterations, and converges to approximately $33\%$ accuracy for large $k$. As expected, larger attack radius breaks the defence, and for $\epsilon=\nicefrac{24}{255}$ the performance of the network is worse than random. These results are consistent with the experiments in \citet{rakin2018parametric} and confirm that noise injection leads to true increased robustness of the network rather than to mere gradient obfuscation \cite{athalye2018obfuscated}.

\section{Conclusions}
\label{sec:conclusion}

In this paper we proposed to inject low-rank colored multi-variate Gaussian noise to the parameters of a CNN during adversarial training. We show that adding covariance terms to the injected noise provides improvement over independent noise \cite{rakin2018parametric} on both white- and black-box attacks. Moreover, even though we used a much smaller architecture (WideResNet-28-4), we achieved results compatible with state-of-the-art adversarial defences, which used WideResNet-34-10. We also performed an ablation study of the method hyperparameter (noise rank) as well as the attack strength ($\epsilon$ and $k$).

%\subsubsection*{Author Contributions}

% The research was funded by the Hyundai Motor Company through the HYUNDAI-TECHNION-KAIST Consortium, the ERC StG RAPID grant, and the Hiroshi Fujiwara Technion Cyber Security Research Center.

{\small
\bibliographystyle{ieee_fullname}
\bibliography{defence_cvpr}
}

\clearpage
\crefalias{section}{appsec}
\crefalias{subsection}{appsec}
\crefalias{subsubsection}{appsec}

\end{document}